\documentclass{ecai} 

\usepackage{latexsym}
\usepackage{amssymb}
\usepackage{amsmath}
\usepackage{amsthm}
\usepackage{booktabs}
\usepackage{enumitem}
\usepackage{graphicx}
\usepackage{color}
\usepackage[colorlinks,linkcolor=red,anchorcolor=magenta,citecolor=blue]{hyperref}
\usepackage{multirow}

\newcommand{\BibTeX}{B\kern-.05em{\sc i\kern-.025em b}\kern-.08em\TeX}

\begin{document}


\begin{frontmatter}


\paperid{957} 


\title{TwinDiffusion: Enhancing Coherence and Efficiency in Panoramic Image Generation with Diffusion Models}


\author[A]{\fnms{Teng}~\snm{Zhou}}
\author[A]{\fnms{Yongchuan}~\snm{Tang}\thanks{Corresponding Author. Email: yctang@zju.edu.cn}}

\address[A]{College of Computer Science and Technology, Zhejiang University}


\begin{abstract}
Diffusion models have emerged as effective tools for generating diverse and high-quality content. However, their capability in high-resolution image generation, particularly for panoramic images, still faces challenges such as visible seams and incoherent transitions. In this paper, we propose TwinDiffusion, an optimized framework designed to address these challenges through two key innovations: the Crop Fusion for quality enhancement and the Cross Sampling for efficiency optimization. We introduce a training-free optimizing stage to refine the similarity of adjacent image areas, as well as an interleaving sampling strategy to yield dynamic patches during the cropping process. A comprehensive evaluation is conducted to compare TwinDiffusion with the prior works, considering factors including coherence, fidelity, compatibility, and efficiency. The results demonstrate the superior performance of our approach in generating seamless and coherent panoramas, setting a new standard in quality and efficiency for panoramic image generation. 
\end{abstract}

\end{frontmatter}


\section{Introduction}
\label{sec:introduction}

\begin{figure*}[!htbp]
    \centering
    \includegraphics[width=1\textwidth]{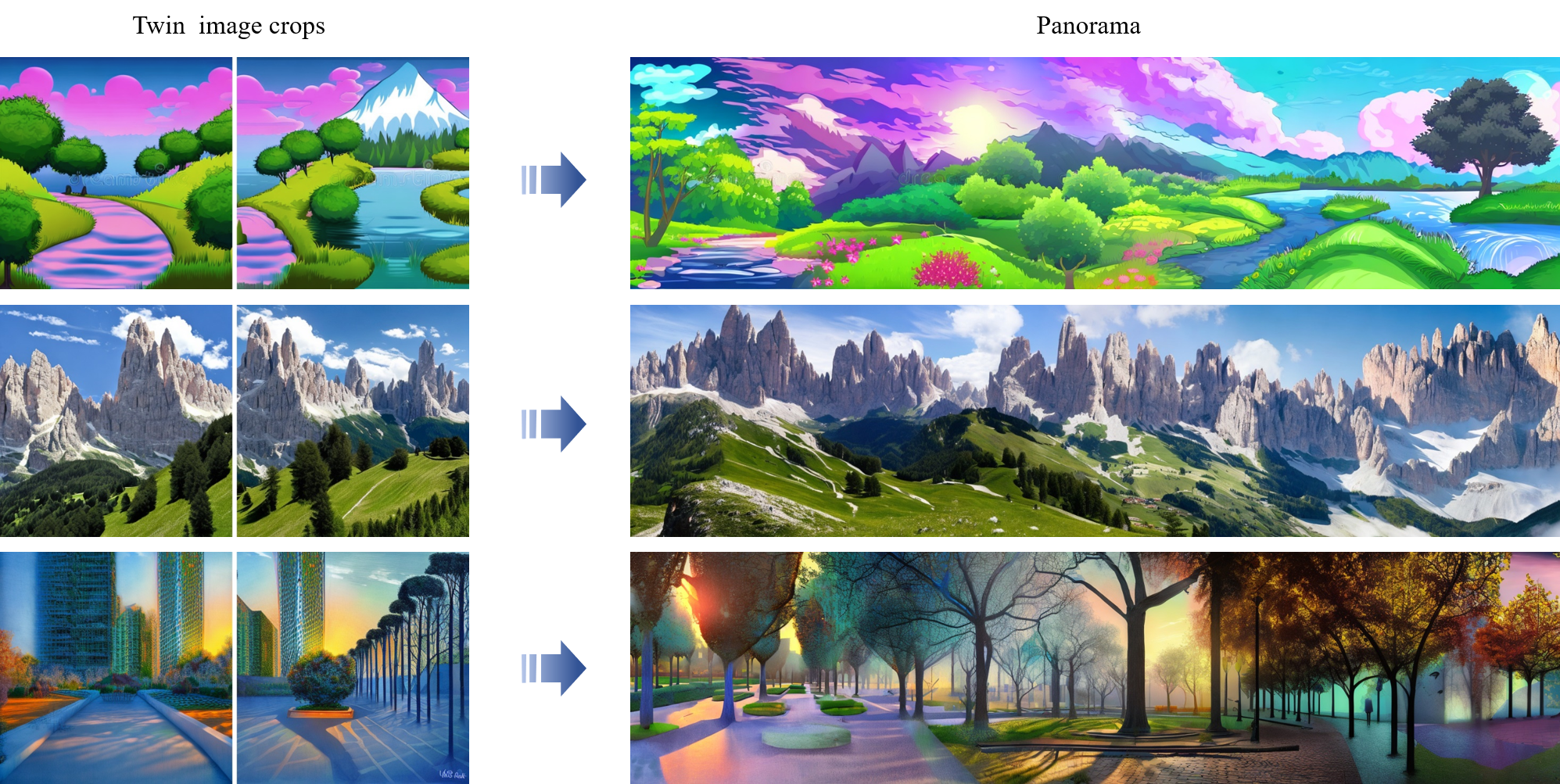}
    \caption{TwinDiffusion is a crop-wise framework designed for high-resolution panorama generation with diffusion models. Inspired by the strong connection between twins, our approach aims to reconcile adjacent areas of the panoramic image space successively. This alignment produces pairs of locally similar image crops resembling twins (left), leading to improved coherence and smoother transitions in panoramas (right).\vspace{1em}}
    \label{fig:intro}
\end{figure*}

Over the past few years, diffusion models \cite{ho2020denoising,rombach2022high,yang2023diffusion} have demonstrated their creativity in generation tasks. They define a pair of forward and reverse Markov chains to learn the data distribution, which bypasses the limitations in other types of generative models like GANs \cite{brock2018large,dhariwal2021diffusion}, VAEs \cite{chira2022image,pandey2021vaes} and Flows \cite{kobyzev2020normalizing}. Diffusion models have risen as effective tools for broad applications, spanning from high-quality images to multi-type content creation. With the growing interest in full information records, immersive virtual reality, artistic expression, and historical preservation, the synthesis of long scrolls is gaining more and more attention, particularly in panoramic image generation tasks \cite{bar2023multidiffusion,feng2023diffusion360,zhang2024taming}.

Recent advancements have exhibited the expansibility of pre-trained diffusion models in generating images of arbitrary dimensions, such as super-resolution diffusion \cite{gao2023implicit,xiao2024single} and area fusion strategies \cite{jimenez2023mixture,xiao2016multi,zhang2023diffcollage}. The latter often involves cropping from a large space into small patches for individual processing, as well as conducting specific guidance to fuse them together, providing more controllability than the former. However, achieving crop-wise high-resolution generation is non-trivial. To our knowledge, MultiDiffusion \cite{bar2023multidiffusion} represents the state-of-the-art framework among the existing methods, yet it still fails to capture the relationships between neighboring image areas, resulting in unnatural connections or even visible seams in panoramas. Although a finer cropping stride could ease such problems, it comes with a higher time cost.

To tackle these challenges, we propose TwinDiffusion, an optimized framework designed to enhance the capability of panoramic image generation with diffusion models. Drawing from the groundwork laid by the MultiDiffusion, our approach introduces two key innovations to make improvements in both quality and efficiency.

\begin{itemize}
\item (\underline{Quality}) \textbf{Crop Fusion}: Our first innovation in TwinDiffusion focuses on refining the coherence of generated panoramas by introducing a training-free optimizing stage. Inspired by the harmonious relationship between twins, this approach is aimed to closely align the adjacent parts of the panoramic image space, leading to smoother transitions and fewer seams in final panoramas as shown in Fig.~\ref{fig:intro}. 

\item (\underline{Efficiency}) \textbf{Cross Sampling}: Our second innovation handles the efficiency of generating panoramas by adopting an interleaving sampling method. With a group of dynamic strides in the cropping process, we effectively mitigate the loss of image quality caused by larger cropping strides, enabling faster generation while upholding the sampling quality.

\item \textbf{Performance Trade-off}: Moreover, we analyze the key factors in TwinDiffusion that impact its performance, including the timestep for introducing the crop optimizing stage, the Lagrange multiplier in our core function, the view stride and the cross stride for sampling image patches. This thorough discussion gives insights into the condition of quality-efficiency balance with our method.
\end{itemize}

Lastly, our comprehensive evaluation of TwinDiffusion compares its performance with baselines in a range of aspects including coherence (measured by LPIPS \cite{zhang2018unreasonable} \& DISTS \cite{ding2020image}), diversity (FID \cite{heusel2017gans} \& IS \cite{salimans2016improved}), compatibility (CLIP \cite{radford2021learning} \& CLIP-aesthetic \cite{schuhmann2022laion}), efficiency (processing time), etc. Qualitatively, we demonstrate its effectiveness and stability in eliminating seams and generating smoother panoramas. Quantitatively, our method outperforms other baselines across all evaluation metrics, striking a new balance in quality and efficiency for panoramic image generation.


\section{Related Work}

\paragraph{Diffusion Models} Diffusion models are inspired by non-equilibrium thermodynamics \cite{sohl2015deep}. They define two Markov chains for forward and reverse processes, namely diffusion and denoising. The forward process is to perturb a data distribution \(x_0\sim q(x)\) into a standard Gaussian distribution \(x_T\sim\mathcal{N}(0,I)\) with \(T\) steps of noise injection. This process is reversed to recreate the sample \(x_0\) that obeys the original data distribution from a Gaussian noise input. Specifically, the reverse process involves training a network to approximate \(q(x_{t-1}\mid x_t)\), and then sampling from \(\mathcal{N}(0,I)\) iteratively with the trained model. From DDPM \cite{ho2020denoising} to DDIM \cite{song2020denoising} to LDM \cite{rombach2022high}, diffusion models have paved the way for text-to-image generation, boosting AI-painting applications like Stable Diffusion \cite{rombach2022high} and DALLE2 \cite{ramesh2022hierarchical}. By capturing the spatial-temporal distribution features, they also show promising prospects in generating multi-modal contents such as videos \cite{blattmann2023align,videoworldsimulators2024,esser2023structure}, audio \cite{huang2023make,liu2023audioldm}, and 3D objects \cite{tang2023make,Wang_2023_CVPR,yao2023dance}.

\paragraph{High-Resolution Image Generation with Diffusion Models} Extensive studies have been dedicated to leveraging diffusion models for controllable high-resolution image generation tasks. The existing methodologies can be divided into two branches: (i) methods that focus on super-resolution \cite{gao2023implicit,xiao2024single} or inpainting \cite{avrahami2023blended} techniques utilizing diffusion processes to infer the missing information, and (ii) methods that center around crop-based fusion strategies within diffusion paths \cite{bar2023multidiffusion,jimenez2023mixture,zhang2023diffcollage}. The former often requires training on specific datasets, combining initial noise with low-resolution images as input to the network. Additionally, they involve resizing the input images, which lacks portability and imposes computational demands. On the other hand, the latter offers greater flexibility by manipulating the generation process in different cropped spaces, and reconciling them in a training-free or fine-tuning manner. Among them, the MultiDiffusion framework proves to be feasible and resultful. However, its optimization function only pays attention to the overlapping regions of image crops, ignoring the non-overlapping adjacent areas, which reduces itself into a naive weighted mean method. Hence, we see room for improvement.

\paragraph{Faster Sampling Method for Diffusion Models} Traditional sampling methods in diffusion require a large number of iterations to generate high-resolution images, which can strain computational resources and slow down operations \cite{franzese2023much,ulhaq2022efficient}. Alongside the scheduler optimization \cite{duan2023optimal,song2020denoising} and diffusion distillation \cite{salimans2022progressive}, we refer to the trajectory stitching and interlaced sampling method \cite{pan2024tstitch} for our sampling acceleration request.


\section{Method}
\label{sec:method}

\begin{figure*}[!htbp]
    \centering
    \includegraphics[width=1\textwidth]{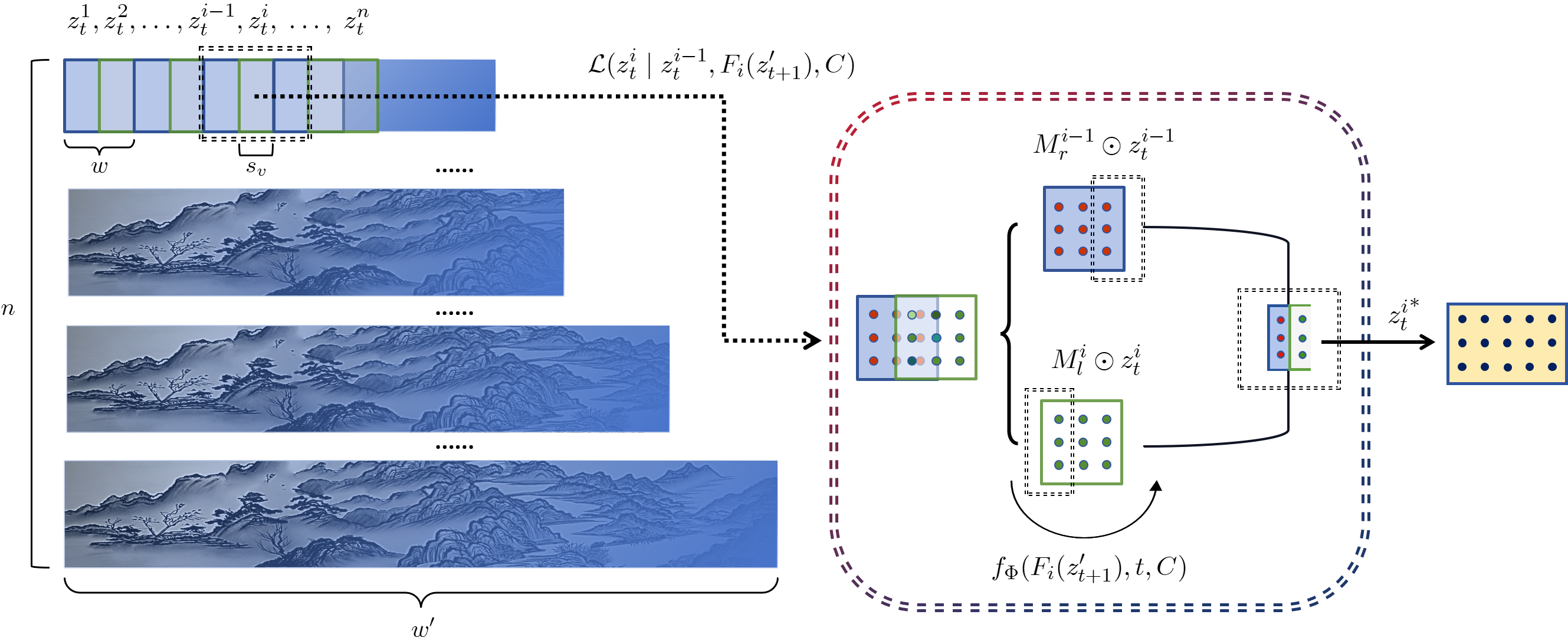}
    \caption{Illustration of our approach applied to panorama generation. The process begins with the mapping function \(F_i\) transforming the image crops into the panoramic space \(Z'\). This results in a sequence of overlapping crops \(z_t^1,z_t^2,\ldots,z_t^n\) arranged spatially, each having an independent denoising path. Our goal is to optimize \(z_t^i\) within the constraints of its adjacent neighbor and itself as well, thus ensuring a unified and progressive fusion of crops. To achieve this alignment, our objective function Eq.~\ref{eq:loss} is defined into two mutual-restricted parts and reaches the minimizer \({z_t^i}^*\) in each denoising timestep: (i) the matching term: differences at the overlaps of \(z_t^i\) and its neighbor \(z_t^{i-1}\), (ii) the regularization term: deviations between \({z_t^i}^*\) and its unoptimized self \(f_\Phi(F_i(z'_{t+1}),t,C)\).\vspace{1em}}
    \label{fig:method}
\end{figure*}

\subsection{Preliminary}
\label{sec:method_0}

To start with, we introduce a pre-trained diffusion model denoted by \(\Phi\), operating in a latent space \(Z=\mathbb{R}^{h\times w\times c}\) and a textual condition \(C\). Employing the deterministic DDIM sampling \cite{song2020denoising}:
\begin{equation}\label{eq:DDIM}
\resizebox{0.9\hsize}{!}{$
    z_{t-1}=\sqrt{\frac{\alpha_{t-1}}{\alpha_t}}z_t+\left(\sqrt{\frac{1}{\alpha_{t-1}}-1}-\sqrt{\frac{1}{\alpha_t}-1}\right)\cdot\Phi(z_t,t,C)
$}
\end{equation}
where \(z_t\in Z\) and \(\alpha_t\) is parameterized by the DDIM schedule \(\{\beta_i\mid i=1,2,\ldots,T,\;\beta_i\in(0,1)\}\), we get image \(z_0\) from initial Gaussian noise \(z_T\) after \(T\) steps of denoising.

Our intention is to extend \(\Phi\) as a reference model to generate images in a larger space \(Z'=\mathbb{R}^{h'\times w'\times c}\), where \(h'>h\) and \(w'>w\). This can be achieved with the MultiDiffusion framework \cite{bar2023multidiffusion}, represented by a function called the MultiDiffuser \(\Psi\). It defines a set of mappings between two model spaces by:
\begin{equation}\label{eq:MD_mapping}
    z^i_t=F_i(z'_t)
\end{equation}
Specifically, \(F_i\) refers to cropping the \(i\)-th image patch from space \(Z'\) with the stride of \(s_v\).

During the MultiDiffusion process, firstly, each crop is simultaneously and independently denoised with \(f_\Phi\): \(z_{t-1}^i=f_\Phi(z_t^i,t,C)\) suggested by Eq.~\ref{eq:DDIM}. Then, based on the Manifold Hypothesis, a least-square optimization for global fusion is formulated to minimize the discrepancy between each crop \(F_i(z'_{t-1})\) and its denoised counterpart \(f_\Phi(F_i(z'_t),t,C)\), merging different crops into one large image \(z'\). According to the properties of \(\Psi\), its optimization problem has an analytical solution. Thus, the minimizer \(z'\) turns out to be a weighted average value:
\begin{equation}\label{eq:MD_ave}
    z'_{t-1}=\frac{\sum_i W_i\odot F_i^{-1}(f_\Phi(z_t^i,t,C))}{\sum_i W_i}
\end{equation}
with \(W_i\) represents the pixel weight matrix of the \(i\)-th crop.

\subsection{Image Crop Fusion}
\label{sec:method_1}

As Eq.~\ref{eq:MD_ave} suggests, the optimization process in MultiDiffusion is only taken for where each \(F_i(z'_t)\) overlaps, disregarding the adjacent but non-overlapping subareas, which can always disrupt the overall coherence of images. The key idea of our Crop Fusion method is to reconstruct this core function, enabling a more reasonable and high-quality panorama generation.

At each denoising timestep \(t\), we get a spatial-ordered sequence of overlapping crops \(z_t^1,z_t^2,\ldots,z_t^n\) generated by the mapping function Eq.~\ref{eq:MD_mapping} in a panoramic space \(Z'\). For each crop \(z_t^i\), our goal is to align its overlapping part of the adjacent crop \(z_t^{i-1}\) as closely as possible, while limiting the deviation from the crop itself. Thus we present:
\begin{equation}\label{eq:loss}
\begin{aligned}
    {z_t^i}^* 
    & =\arg\min_{z_t^i\in Z'}\mathcal{L}(z_t^i\mid z_t^{i-1},F_i(z'_{t+1}),C) \\
    & =\arg\min_{z_t^i\in Z'}\parallel M_r^{i-1}\odot z_t^{i-1}-M_l^i\odot z_t^{i}\parallel^2+ \\
    & \quad\quad\quad\quad\quad\;\parallel f_\Phi(F_i(z'_{t+1}),t,C)-z_t^i\parallel^2
\end{aligned}
\end{equation}
as our optimization task, where \(z^*\) denotes the optimized crop, \(M_l,M_r\in\{0,1\}^{h\times w}\) represent the binary masks covering the crop's left and right overlapping regions according to \(s_v\), and \(\odot\) is the Hadamard product. The second part of Eq.~\ref{eq:loss} serves as a regularization term, which is used to coordinate the alignment behavior of crops. Therefore, the objective function of TwinDiffusion can be formulated as follows:
\begin{equation}\label{eq:loss_st}
\begin{aligned}
    \text{min}\quad & \parallel M_r^{i-1}\odot z_t^{i-1}-M_l^i\odot z_t^{i}\parallel^2 \\
    \text{s.t.}\quad & \parallel f_\Phi(F_i(z'_{t+1}),t,C)-z_t^i\parallel^2\enspace\leq\;\aleph
\end{aligned}
\end{equation}
and the Lagrangian function for this problem is given by:
\begin{equation}\label{eq:loss_lagr}
\begin{aligned}
    L(z_t^i,\lambda)
    & =\parallel M_r^{i-1}\odot z_t^{i-1}-M_l^i\odot z_t^{i}\parallel^2+ \\
    & \lambda(\parallel f_\Phi(F_i(z'_{t+1}),t,C)-z_t^i\parallel^2-\;\aleph)
\end{aligned}
\end{equation}
where \(\lambda\) is the Lagrange multiplier associated with the constraints of adjacent but non-overlapping regions in the panorama. Using the Karush-Kuhn-Tucker (KKT) conditions, we can reach an optimal solution:
\begin{equation}\label{eq:loss_solu}
    {z_t^i}^*=
    \begin{cases}
        M_l^i\odot {z_t^i}^*=(1+\lambda)^{-1}[M_r^{i-1}\odot z_t^{i-1}+\\ 
        \quad\qquad\qquad\qquad\qquad\lambda M_l^i\odot f_\Phi(F_i(z'_{t+1}),t,C)] \\[0.5em]
        M_r^i\odot {z_t^i}^*=M_r^i\odot f_\Phi(F_i(z'_{t+1}),t,C)
    \end{cases}
\end{equation}
which demonstrates that our Crop Fusion is a training-free method with closed-form optimization.

As depicted in Fig.~\ref{fig:method}, our framework progressively optimizes image crops while considering their coherence in multiple subregions and achieving a unified fusion. This approach fundamentally differs from MultiDiffusion, which performs a single weighted average across the entire panorama space.

\subsection{Cross Sampling}
\label{sec:method_2}

In Eq.~\ref{eq:MD_mapping}, \(F_i\) specifies a sliding window to crop overlapping images with a fixed step size \(s_v\), referred to as the view stride later. We notice that the quality and generation speed of panoramic images heavily depend on this view stride, which determines the degree of overlap between crops. A finer degree of overlap results in superior panoramas, yet processing numerous image crops during denoising iterations can be time-consuming.

To address this quality-efficiency trade-off, we propose a variant mapping function called Cross Sampling defined by:
\begin{equation}\label{eq:cross_mapping}
    z_t^i=F_i^{(k)}(z_t'),\quad k=t\!\!\!\mod\:r
\end{equation}
where \(r\) controls the interleaving frequency and \(k\) denotes the sampling mode. Staggering in \(r\) times, our method dynamically forms a set \(\mathcal{Z} =\{z_{i,j}\mid1\leq i\leq r, 1\leq j\leq n\}\) consists of \(r\) groups, each containing \(n\) overlapping crops, with spatial locations incrementing by the cross stride \(s_r\). Then, in \(T\) rounds of denoising, we alternate between using these crop groups for sampling in \(r\) staggered spaces.

Panorama seams mostly occur where the crops meet. Thus by incorporating the Cross Sampling method, TwinDiffusion takes more flexible control over the fine degree of overlap between neighboring image crops, filling the gaps caused by the enlarged view stride. Our experimental results in Sec.~\ref{par:sv_abl} have demonstrated that with this straightforward yet effective solution, we can double \(s_v\) to cut the generation time in half or more (with a larger \(s_v\)), while maintaining panoramic image quality on par with the original MultiDiffusion.


\section{Experiments}

\subsection{Panorama with Twin Crops}

Here, we report two successive implementations of TwinDiffusion: the Single form, generating twin images with high and controllable similarity, and the Multiple form, extending its capabilities to synthesize panoramas composed of multiple, harmoniously interconnected twin crops. 

\begin{figure}[!htbp]
    \centering
    \includegraphics[width=1\linewidth]{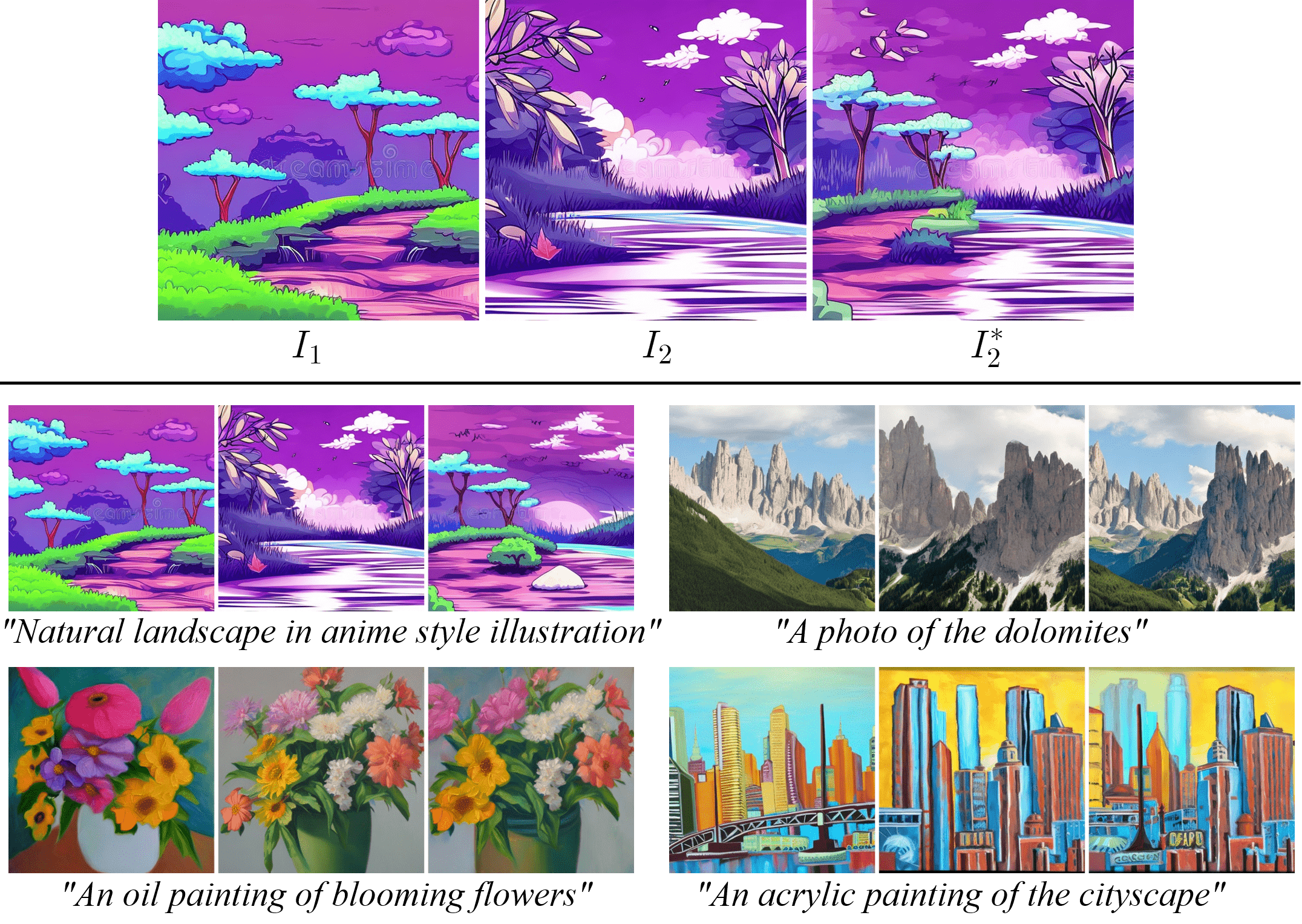}
    \caption{Applying our Crop Fusion method to generate twin images. Top: the optimized \(I_2^*\) exhibits a seamless fusion effect that meets our expectations. Bottom: We further test its limits by fixing the regularization term of Eq.~\ref{eq:loss_st}. The results demonstrate our method's robustness under extreme conditions.\vspace{1.2em}}
    \label{fig:twin_ex}
\end{figure}

\paragraph{Single: Twin Images} As illustrated in Fig.~\ref{fig:twin_ex} (top), our method generates a pair of images that locally resemble each other like twins.\(I_1\) and \(I_2\) represent the first and second images respectively, generated from initial latent noise satisfying \(M_r^1\odot z_1=M_l^2\odot z_2\). \(I_2^*\) corresponds to \(I_2\) with our optimization. It successfully retains the content of original \([I_2]_r\) while closely aligning its left part to \([I_1]_r\), achieving a seamless fusion of \([I_1]_r\) and \([I_2]_l\). 

We also conduct a stress test on TwinDiffusion's ability to fuse image crops. Specifically, we replace \(f_\Phi(F_i(z'_{t+1}),t,C)\) in the regularization term of Eq.~\ref{eq:loss_st} with a constant reference \(\tilde{z}_t^i\), representing the raw \(z_t^i\) following its unoptimized denoising trajectory. The results in Fig.~\ref{fig:twin_ex} (bottom) demonstrate that even under such conditions, our method still achieves a natural and seamless crop fusion. This work serves as an initial validation to ensure higher consistency in the subsequent panorama generation.

\paragraph{Multiple: Panorama with Twin Crops} Beyond a single pair of images, we generalize this approach to a sequence of images, i.e., crops \(z_t^1,z_t^2,\ldots,z_t^n\) within panoramas. Our optimization is applied to each pair of neighboring crops, promoting a high degree of similarity between \([z_t^i]_r\) and \([z_t^{i+1}]_l\), thereby resulting in higher-quality panoramas with consecutive twin crops as depicted in Fig.~\ref{fig:intro}. Refer to Appendix~\ref{appendix:twin_img} for more implementation examples.

\subsection{Comparison}
\label{sec:comparison}

We conduct a comprehensive evaluation of our approach from both qualitative and quantitative perspectives, comparing images generated by TwinDiffusion versus other baselines.

For the reference model \(\Phi\), we employ two variants: the widely used diffusion model Stable Diffusion v2.0, and its advanced version Stable Diffusion XL v1.0 \cite{podell2023sdxl}. They respectively operate in an image space of \(\mathbb{R}^{512\times512\times3}\) and \(\mathbb{R}^{1024\times1024\times3}\). We align the size of crops with the default resolution of \(\Phi\), creating panoramas in \({512\times2048}\) and \({1024\times4096}\) correspondingly. To ensure the reliability of our results, we test 20 different prompts involving various contents and art styles, and generate 200 panoramas per prompt with 5 sets of random seeds. The results presented in Sec.~\ref{sec:comparison} are specifically obtained with a reference model of \(\Phi=\text{SD}_{\text{2.0}}\), a crop fusion timestep of \(\tau=T/2\), an adjacent control factor of \(\lambda=1\), a view stride of \(s_v=16\) and a cross stride of \(s_r=8\). More details and comparisons about these contributing factors are thoroughly discussed in Sec.~\ref{sec:ablation} and Appendix~\ref{appendix:ref_models}.

\begin{figure*}[!htbp]
    \centering
    \includegraphics[width=1\textwidth]{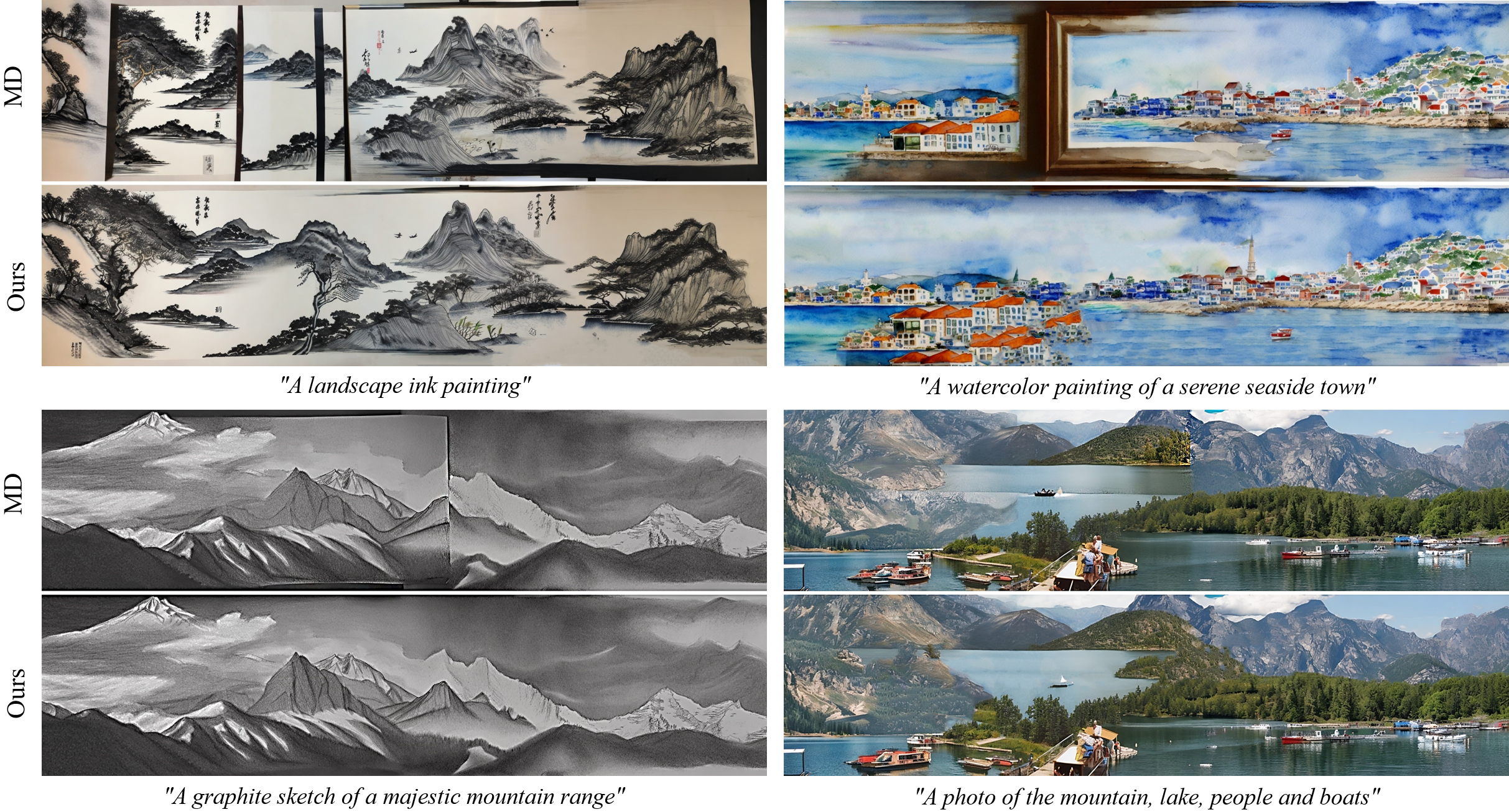}
    \caption{Qualitative comparisons between MultiDiffusion and ours. Our approach significantly reduces the odd joints and visible seams that commonly occur in MultiDiffusion, resulting in higher-quality panoramic images.\vspace{1em}}
    \label{fig:qual_comp}
\end{figure*}

\paragraph{Qualitative Comparison} In Fig.~\ref{fig:qual_comp}, we showcase the comparative performance between our method and MultiDiffusion across a series of qualitative examples. Our TwinDiffusion effectively mitigates the problem of visible seams at the overlaps of image crops, achieving a smoother transition where MultiDiffusion tends to struggle. As shown in the first three cases, this improvement is particularly noticeable for art painting, which differs from the natural landscape due to its frame-like incoherence that often arises at the edges of crops. More qualitative results are provided in Appendix~\ref{appendix:qual_comp}.

\paragraph{Quantitative Comparison} We utilize a range of quantitative metrics focusing on the following four aspects: (i) coherence at the intersection of crops, (ii) fidelity and diversity of the generated panoramas, (iii) compatibility with the input prompts, as well as (ix) efficiency of the optimization process. Recognizing that resizing the entire panoramic image to meet the small dimensions required by the metrics (e.g. \(299^2\) for FID, \(224^2\) for CLIP) could lead to loss of essential features and distortions, we choose to test with images cropped from panoramas at a \(512^2\) resolution instead. 

\begin{itemize}
    \item (\underline{Coherence}) \textbf{Learned Perceptual Image Patch Similarity (LPIPS)} and \textbf{Deep Image Structure and Texture Similarity (DISTS)}: LPIPS and DISTS capture the perceptual differences between two images by computing distances of their feature vectors. Each generated panorama is divided into 8 pairs of adjacent but non-overlapping image crops according to the \(s_v\). From these cropped views, we randomly take 4,000 pairs to compute LPIPS and DISTS values. 
    
    \item (\underline{Fidelity \& Diversity}) \textbf{Fréchet Inception Distance (FID)} and \textbf{Inception Score (IS)}: Leveraging the underlying output of Inception V3 network, FID and IS describe both fidelity and diversity of generated images. Inspired by \cite{bar2023multidiffusion}, we measure FID and IS between the distribution of generated and reference image sets, where the former consists of images cropped from panoramas, and the latter comprises images generated by reference models. To avoid coherence interfering with diversity, we extract only one random crop from each panorama, and calculate the metrics from each crop to the reference image set.
    
    \item (\underline{Compatibility}) \textbf{CLIP} and \textbf{CLIP-aesthetic}: CLIP is used to assess the cosine similarity between generated images and the input prompts, while CLIP-aesthetic score is predicted from a linear estimator on top of CLIP. For a given prompt, we employ the cropped image set mentioned above to compute the scores.

    \item (\underline{Efficiency}) \textbf{Generation Time}: The time taken to generate a complete panoramic image is evaluated on an A100 GPU.
\end{itemize}

We make comparisons among the Blended Latent Diffusion \cite{avrahami2023blended}, MultiDiffusion and our TwinDiffusion, calculating the mean and standard deviation scores of all metrics. Additionally, we measure the performance of \(\Phi\) itself (i.e. the Stable Diffusion), which is measured by internal comparisons within the reference image set.

\begin{figure}[!htbp]
    \centering
    \includegraphics[width=1\linewidth]{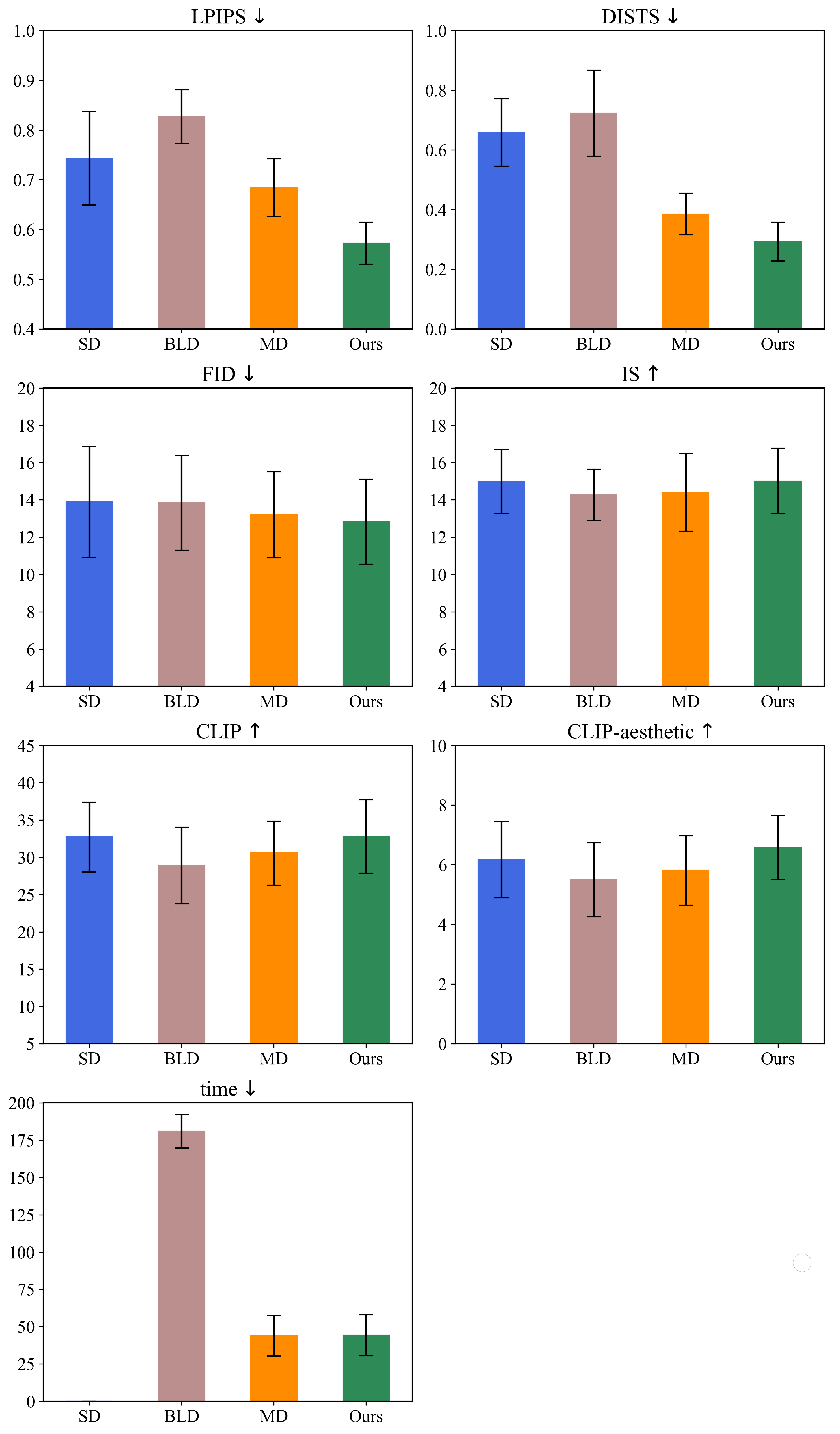}
    \caption{Quantitative results comparing our approach with the baselines. Our method demonstrates its superiority in all aspects, particularly in coherence, the best visual representation of panoramas.\vspace{1.5em}}
    \label{fig:quant_comp}
\end{figure}

As reflected in Fig.~\ref{fig:quant_comp}, our approach stands out as the optimal method across all evaluation criteria. Coherence, the most important aspect of panoramic images, is greatly improved by our method, as reported in the first row. Additional progress can also be observed in the CLIP-aesthetic metric, implying that the increase in coherence could facilitate the compatibility and aesthetic appeal of generated results. Meanwhile, we get comparable scores in FID and IS. This keeps in line with our method's primary focus on the seam issue, which may not have much impact on fidelity and diversity. In terms of efficiency, our method achieves better image quality without any compromise in time cost.

\subsection{Ablation}
\label{sec:ablation}

TwinDiffusion incorporates several key factors that contribute to its performance, including the timestep \(\tau\) for introducing the Crop Fusion stage, the adjacent control factor \(\lambda\) in our optimization function, the view stride \(s_v\) for cropping image patches, and the cross stride \(s_r\) in the Cross Sampling method.

\begin{figure}[!htbp]
    \centering
    \includegraphics[width=1\linewidth]{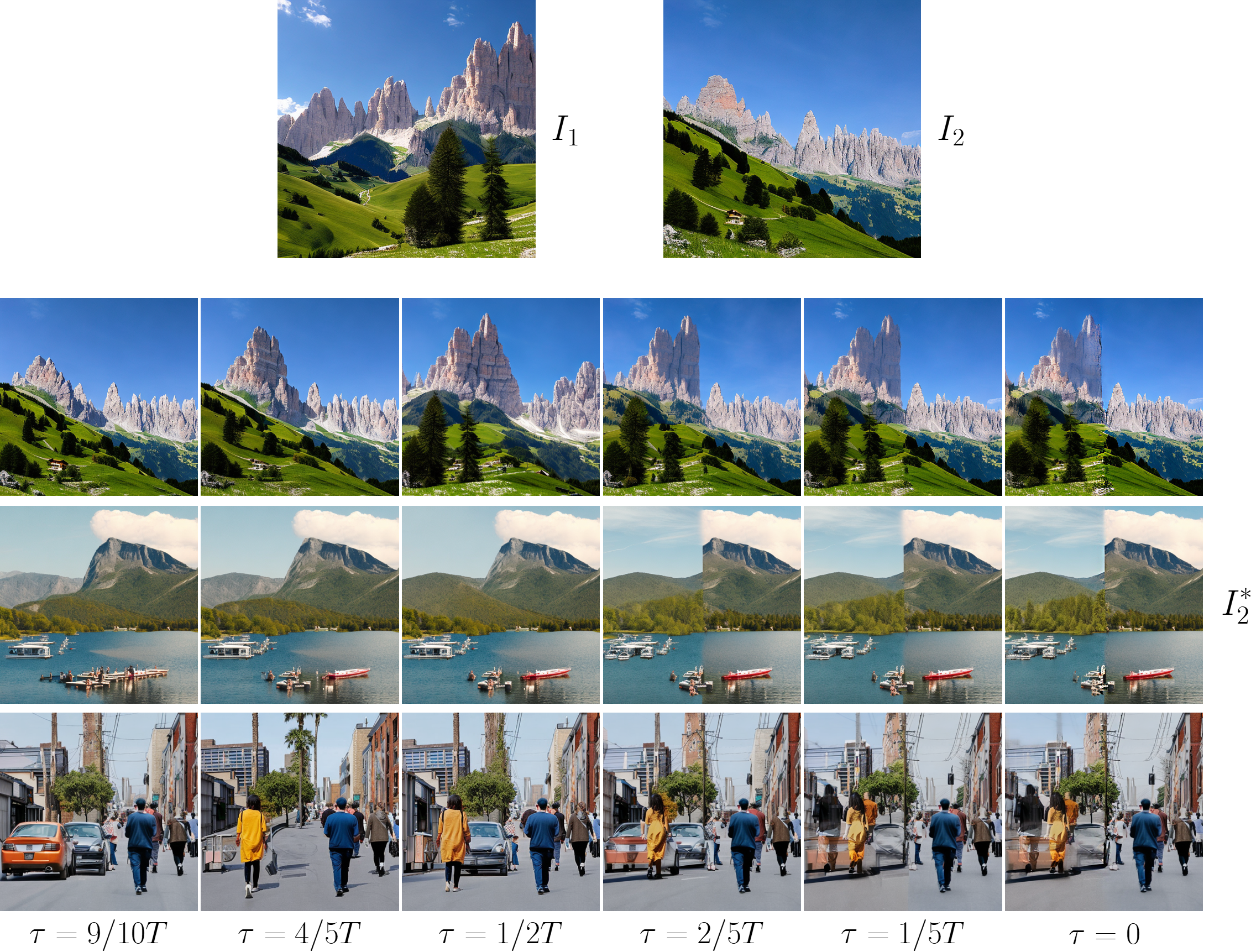}
    \caption{The analysis of the proper timestep \(\tau\) for introducing the Crop Fusion stage. As \(\tau\) decreases, a gradual transition from under-optimization to over-optimization can be observed, with the best results attained by \(\tau=T/2\). The bottom rows of the figure offer two additional examples that further support our findings.\vspace{1.25em}}
    \label{fig:optimization_time_abl}
\end{figure}

\paragraph{Optimization Timestep} In the initial attempts at twin-image generation, we apply the Crop Fusion method throughout the entire denoising process. However, the outcomes are unsatisfactory as the generated images exhibit a distinct left-right fragmentation, where an appropriate optimization time window is needed to achieve the desirable results. We know that an earlier guidance plays a greater role in the diffusion trajectory. Thus, we decide to confine the optimization period to the early stages of the sampling process. That is, the Crop Fusion is carried out from \(t\): \(T\rightarrow{\tau}\) and then stopped during \(t\): \(\tau\rightarrow0\), where \(T\) represents the total timestep of the corresponding diffusion scheduler. 

The relationship between \(\tau\) and the stitching effect of twin images is depicted in Fig.~\ref{fig:optimization_time_abl}. We can see a progressive effect on \(I_2^*\) with the decrease of optimization timestep \(\tau\). (i) When \(\tau>T/2\), it is under-optimized, as \([I_2^*]_l\) differs significantly from \([I_1]_r\). (ii) When \(\tau=T/2\), it is well-optimized, generating the most natural and seamless \(I_2^*\). (iii) When \(\tau<T/2\), it is over-optimized, \(I_2^*\) stays too close with \([I_1]_r\) while failing to fuse with \([I_2^*]_r\).

\begin{figure}[!htbp]
    \centering
    \includegraphics[width=1\linewidth]{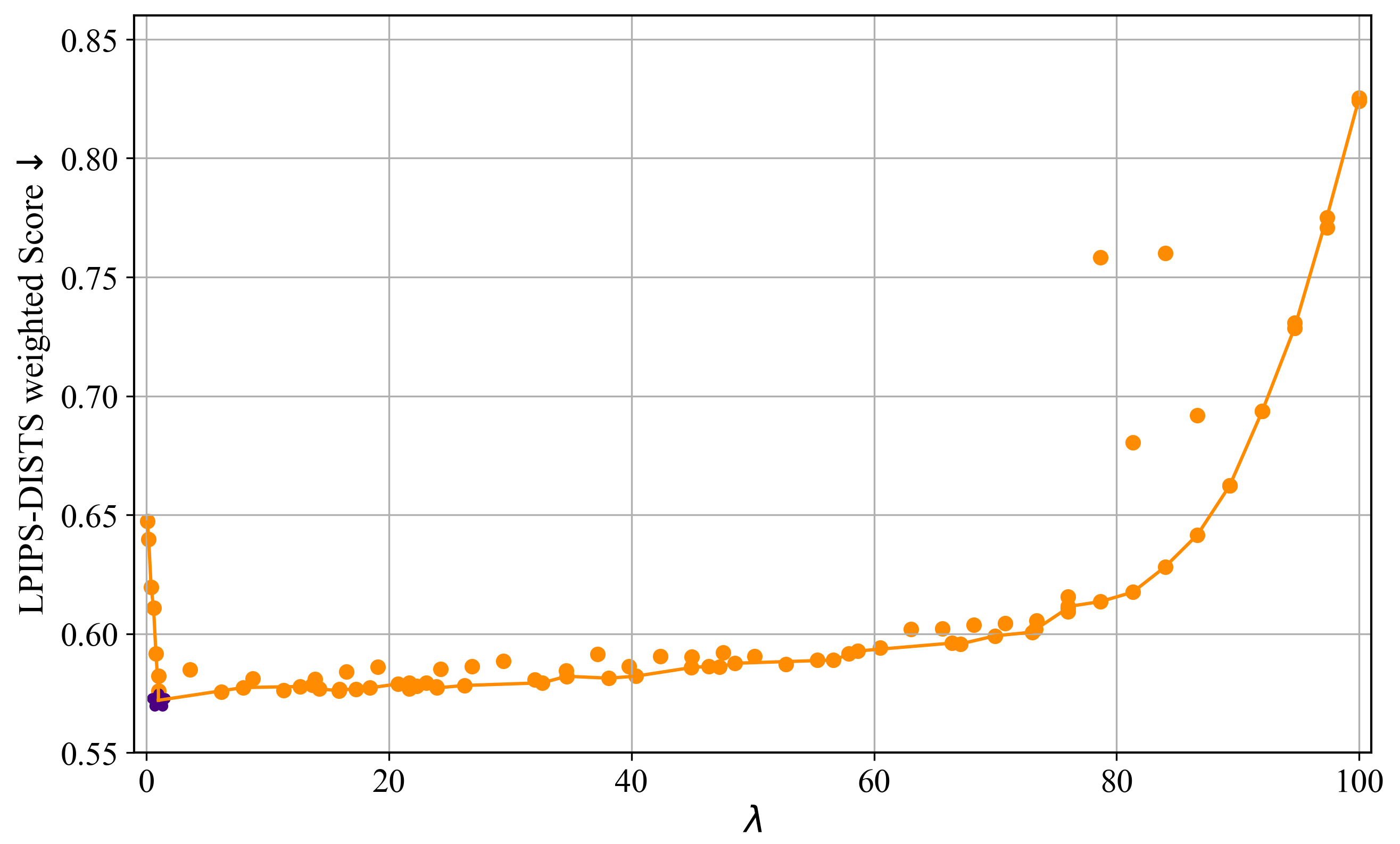}
    \caption{Further explorations about adjacent control factor \(\lambda\). It presents a comparative analysis of different \(\lambda\) values to study their impact on the alignment behavior and visual coherence of panoramas. The results demonstrate that our method achieves the best balance around \(\lambda=1\) and is not sensitive to changes of \(\lambda\) values.\vspace{1.2em}}
    \label{fig:lambda_abl}
\end{figure}

\paragraph{Adjacent Control} The control factor \(\lambda\) in Eq.~\ref{eq:loss_lagr} plays a crucial role in determining the alignment behavior of image crops. It allows us to adjust the balance between aligning closely with adjacent blocks and maintaining self-alignment, thus significantly influencing the overall quality of panoramic images. Since coherence is an essential attribute in panoramas, we streamline our assessment to focus on LPIPS and DISTS metrics to represent image quality.

\begin{figure}[!htbp]
    \centering
    \includegraphics[width=1\linewidth]{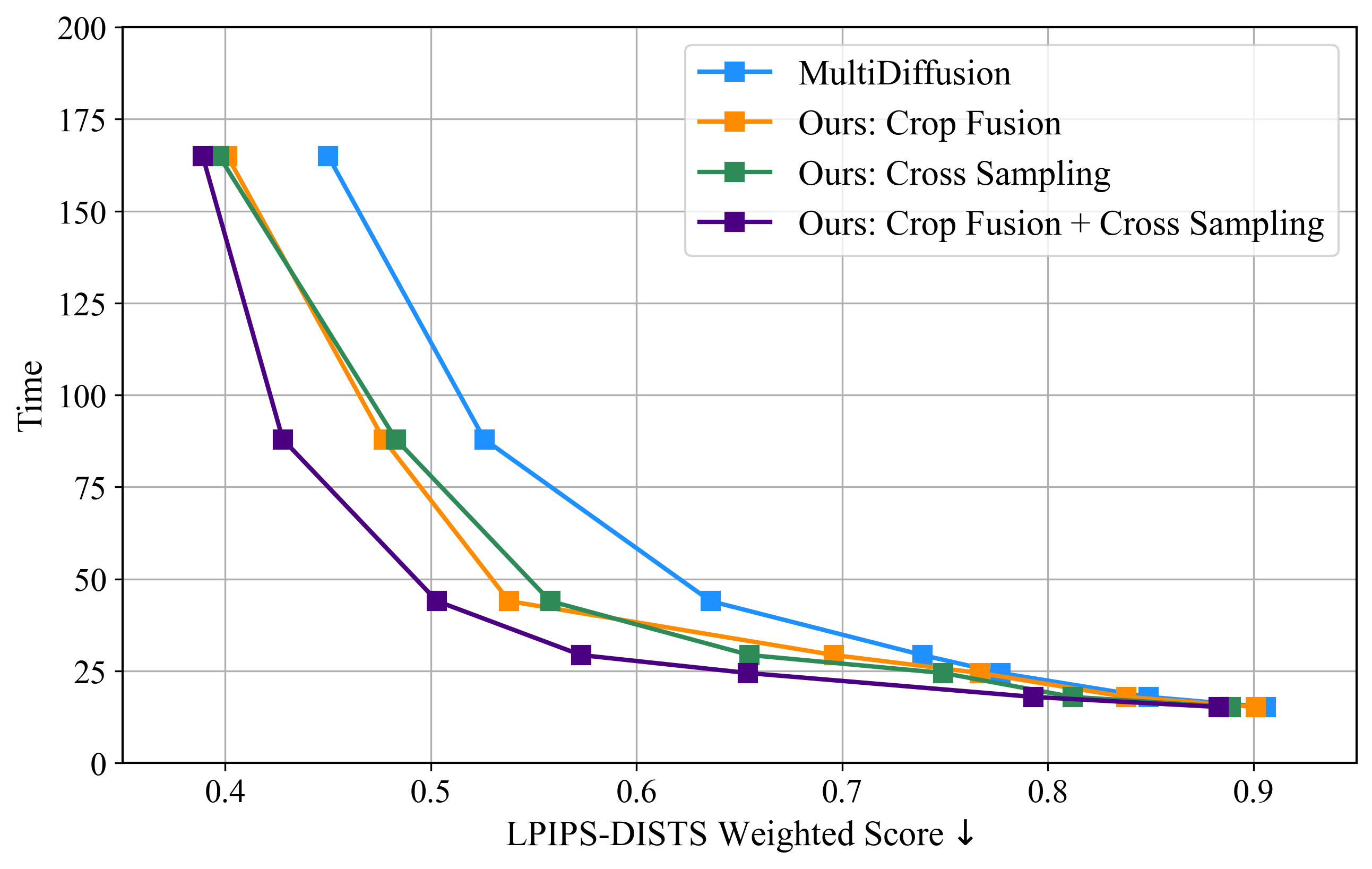}
    \caption{Ablation study about our method in improving the quality-speed trade-off. As seen, the effect of the single Crop Fusion and Cross Sampling method is comparable, and they both outperform the MultiDiffusion baseline. With their additive effects, TwinDiffusion exhibits a more robust rate of curve, demonstrating a better speed and quality balance.\vspace{1.5em}}
    \label{fig:sv_abl}
\end{figure}

In Fig.~\ref{fig:lambda_abl}, we compare the results across a wide range of \(\lambda\) values from 0.1 to 100.  When \(\lambda\) is between 0.1 and 80, the LPIPS-DISTS weighted score shows a very slow growth trend. It only starts to increase greatly when \(\lambda\) exceeds 80. The lowest point is concentrated around 1, where our method achieves the optimal balance between the two competing terms mentioned above, leading to the desired visual consistency in panoramic images. This is reasonable and in keeping well with our objective.

\paragraph{View Stride}\label{par:sv_abl} As discussed in Sce.~\ref{sec:introduction} and Sec.~\ref{sec:method_2}, the stride of neighboring views controls the trade-off between quality and efficiency. Smaller \(s_v\) results in better image quality but also takes a lot of time; larger \(s_v\) can accelerate the generation speed of images but the overall quality falls. To test the effectiveness of our approach in improving this problem, we measure the generation time and the aforementioned LPIPS-DISTS weighted score of generated images under \(s_v=4,8,16,24,32,40,48\), making a comparison between MultiDiffusion and our TwinDiffusion with and without the two optimization method.

\begin{table*}[!htbp]
	\centering
	\caption{Here are the findings from our further exploration of the Cross Sampling strategy. Different \(s_r\) values were tested across all metrics. The better effects are more likely to be achieved when \(s_r=s_v/2,\,s_v/3\), indicating that excessively fine cross strides are unnecessary.\vspace{2em}}
	\label{tab:sc_abl}
    \resizebox{0.9\textwidth}{!}{
	\begin{tabular}{cccccccc}
        \hline
		\multirow{2}{*}{} & \multicolumn{2}{c}{Coherence} & \multicolumn{2}{c}{Fidelity \& Diversity} & \multicolumn{2}{c}{Compatibility} & Efficiency \\
        \cmidrule(lr){2-3} \cmidrule(lr){4-5} \cmidrule(lr){6-7} \cmidrule(lr){8-8}  
		& LPIPS\(\downarrow\) & DISTS\(\downarrow\) & FID\(\downarrow\) & IS\(\uparrow\) & CLIP\(\uparrow\) & CLIP-aesthetic\(\uparrow\) & time\(\downarrow\) \\ 
		\hline
		w/o Cross Sampling & 0.69 & 0.49 & \underline{14.03} & 12.31 & 30.91 & 6.43 & \textbf{43.98} \\
		\(s_r=s_v/2\) & \textbf{0.60} & \underline{0.43} & \textbf{13.16} & \textbf{14.49} & \textbf{32.06} & \textbf{6.76} & \underline{43.99} \\
		\(s_r=s_v/3\) & \underline{0.63} & \textbf{0.42} & 14.29 & \underline{14.15} & \underline{31.38} & 6.46 & 45.01 \\
        \(s_r=s_v/5\) & 0.71 & 0.59 & 15.11 & 13.98 & 30.93 & 6.42 & 47.52 \\
        \(s_r=s_v/7\) & 0.76 & 0.62 & 16.00 & 13.98 & 30.25 & \underline{6.48} & 53.60\\
        \(s_r=s_v/s_v\) & 0.84 & 0.78 & 16.21 & 12.58 & 30.26 & 6.05 & 59.67 \\
		\hline
	\end{tabular}}
\end{table*}

As seen in Fig.~\ref{fig:sv_abl}, our method surpasses the baseline in a better quality-efficiency balance. (i) For quality: under the same view stride, both the Crop Fusion and Cross Sampling methods reach a lower LPIPS-DISTS weighted score than the MultiDiffusion, and full TwinDiffusion reaches even lower scores due to the combined effects of the two. (ii) For efficiency: analyzing the blue and green lines, we can observe that our Cross Sampling approach successfully enables generating comparable results within a fraction of the time required by the original MultiDiffusion, specifically achieving a reduction of N-fold when using a \(N\times s_v\) value. (iii) For quality-efficiency trade-off: the slopes indicate that our method takes greater advantage of the efficiency gained from larger \(s_v\) while maintaining high image quality, striking an optimal balance between quality and speed for panoramic image generation.

\paragraph{Cross Stride} We further investigate the influence of cross stride to ensure an appropriate interleaving frequency in the Cross Sampling method. In particular, \(s_r=1\) means setting a different sampling mode per pixel, and \(s_r=s_v\) is equivalent to not using our Cross Sampling method. All the results are obtained under the same condition described in Sec.~\ref{sec:comparison}.

Tab.~\ref{tab:sc_abl} provides a series of comparisons on all sides, with the best and the second-best results marked in bold and underlined respectively. The scores show a stable and consistent pattern, where the most desirable outcomes are generally achieved when \(s_r\) is set to \(s_v/2\) or \(s_v/3\). In the extreme case of \(s_r=s_v/s_v=1\), the effect regresses to no Cross Sampling or even worse. This observation suggests that a finer interleaving level of sampling does not contribute to the quality scores but adds unnecessary running time.


\section{Conclusion}

In this paper, we have presented TwinDiffusion, an optimized framework for panoramic image generation using state-of-the-art diffusion models. Our work breaks through the existing limitations in quality and efficiency by introducing two key innovations: (i) a lightweight fusion stage to enhance coherence, and (ii) an interleaved sampling method to improve generating speed. By extending this promising framework to wider domains, especially virtual reality and graphic design, we can unlock new possibilities for creating dynamic and immersive visual content. 

\paragraph{Limitations and Social Impact} Although TwinDiffusion works well in most cases, it still faces some limitations. Our approach primarily focuses on optimizing the local similarity of the image areas. However, it could not ensure stability in perceiving the overall layout of the images, which may lead to the generation of visually coherent but spatially illogical panoramas. As for potential negative impact, image generation models may involve personal copyright or generate fake, offensive, discriminatory results. Further research should prioritize the responsible use of the relevant technology to avoid generating content in any harmful way.



\begin{ack}
We would like to express sincere gratitude to Xiaoyu Zhang, Mingyue Hu, and the entire research group for their inspiration and support. Furthermore, we thank Yunhao Chen for assisting in diagnosing issues and providing crucial feedback and suggestions.
\end{ack}



\bibliography{mybibfile}


\clearpage
\appendix
\onecolumn

\setcounter{figure}{0}
\setcounter{table}{0}
\renewcommand\thefigure{A\arabic{figure}}
\renewcommand\thetable{A\arabic{table}}

\section{More Implementation Examples on Twin-Image Generation}
\label{appendix:twin_img}

\begin{figure*}[!htbp]
    \centering
    \includegraphics[width=1\linewidth]{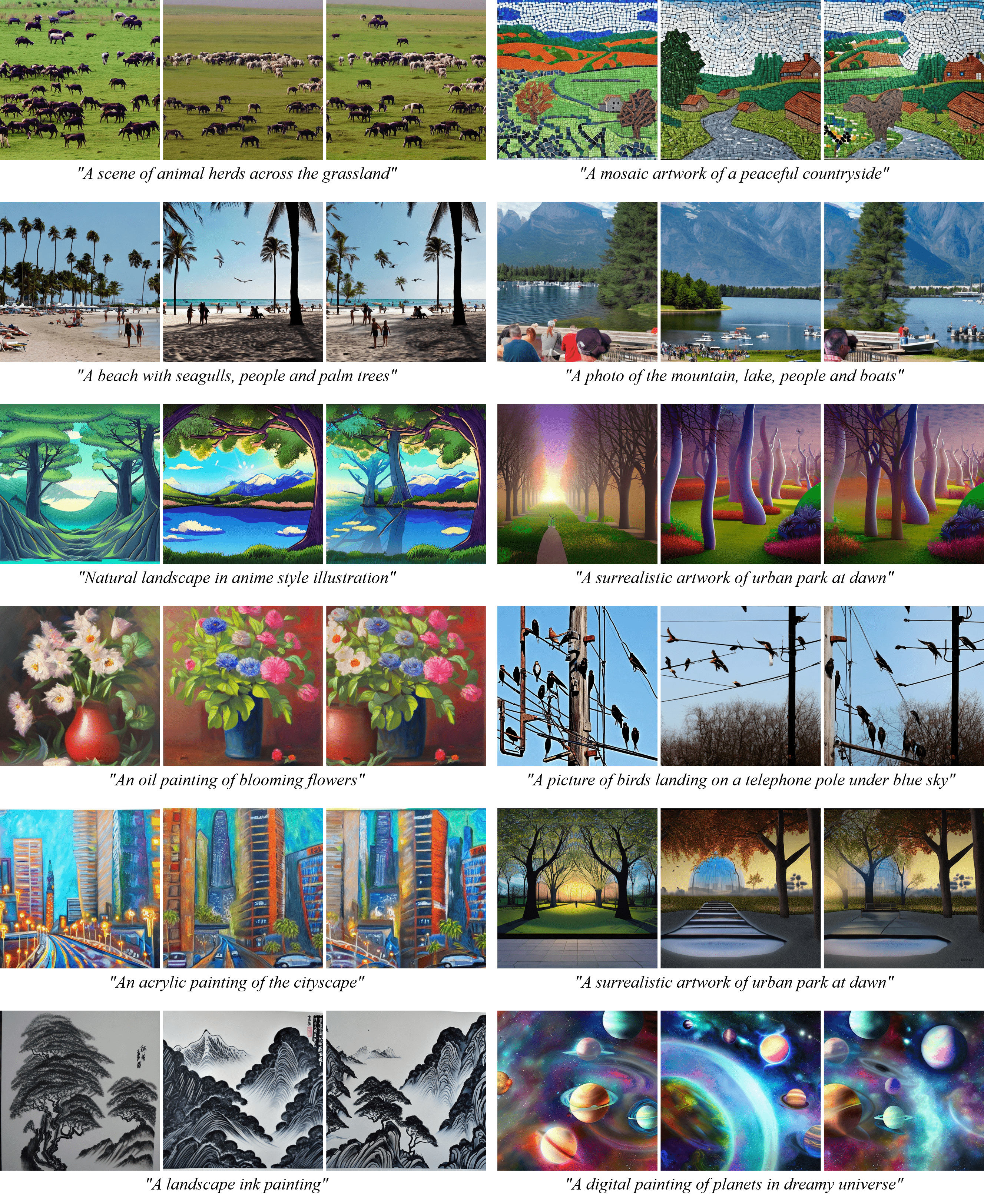}
    \caption{More examples of twin images with our TwinDiffusion.\vspace{1em}}
    \label{fig:twin_ex_plus}
\end{figure*}

Our experimental results of twin images are provided in Fig.~\ref{fig:twin_ex_plus}, meaning the high stability of our method.

\section{More Results of Qualitative Comparison on Panorama Generation}
\label{appendix:qual_comp}

\begin{figure*}[!htbp]
    \centering
    \includegraphics[width=1\linewidth]{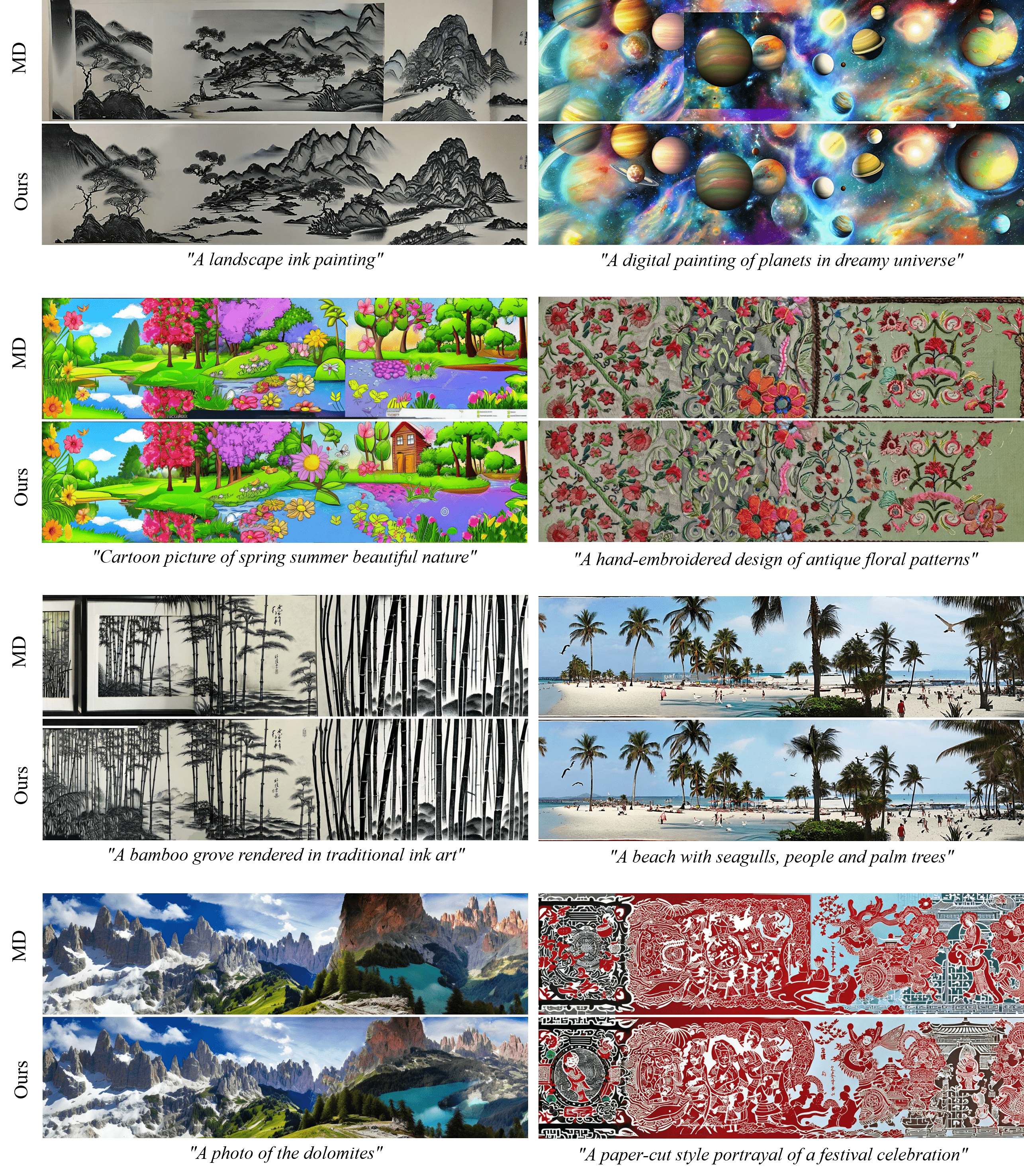}
    \caption{Additional qualitative comparison results on panorama generation with various text prompts.\vspace{1em}}
    \label{fig:qual_comp_plus}
\end{figure*}

Fig.~\ref{fig:qual_comp_plus} gives more qualitative comparisons with MultiDiffusion on panoramic image generation, highlighting the areas where improvements have been made.

\section{Analysis on Different Reference Models}
\label{appendix:ref_models}

In addition to Stable Diffusion v2.0, we also conduct a series of experiments on Stable Diffusion XL v1.0 as our reference model \(\Phi\), aiming to further explore the generalization capability of our approach. Considering that images generated by SDXL have larger dimensions of \(1024^2\), we also crop them into \(512^2\), which aligns with the other image sets.

Tab.~\ref{tab:rm_ablation} presents a comprehensive comparison among the methods using different reference models. As observed, even in the context of SDXL, our approach still contributes to refining the visual and semantic coherency of generated panoramas, albeit to a lesser extent than observed with the SD-based reference model. This observation is understandable, considering the inherent attributes of SDXL as an advanced diffusion model that doubles the default resolution in both height and width. Despite the limitations imposed by the reference model, our approach still enhances several metrics in panoramic image generation tasks, demonstrating its ability to improve panorama quality.

\begin{table}[!htbp]
	\centering
	\caption{Comparisons among different methods using various reference models. Despite the advanced performance stored in SDXL itself, our approach still demonstrates its ability to improve the quality and coherence of the generated panoramas.\vspace{2em}}
	\label{tab:rm_ablation}
 	\vspace{5pt}
    \resizebox{0.9\textwidth}{!}{
	\begin{tabular}{cccccccc}
        \hline
		\multirow{2}{*}{} & \multicolumn{2}{c}{Coherence} & \multicolumn{2}{c}{Fidelity \& Diversity} & \multicolumn{2}{c}{Compatibility} & Efficiency \\
        \cmidrule(lr){2-3} \cmidrule(lr){4-5} \cmidrule(lr){6-7} \cmidrule(lr){8-8}  
		& LPIPS\(\downarrow\) & DISTS\(\downarrow\) & FID\(\downarrow\) & IS\(\uparrow\) & CLIP\(\uparrow\) & CLIP-aesthetic\(\uparrow\) & time\(\downarrow\) \\ 
		\hline
        \(\text{SD}_{\text{2.0}}\) & 0.74 & 0.66 & 13.88 & 14.98 & 32.43 & 6.18 & / \\
		MD\(_{\Phi=\text{SD}_{\text{2.0}}}\) & 0.68 & 0.39 & 13.46 & 13.20 & 31.55 & 5.81 & 43.92 \\
		Ours\(_{\Phi=\text{SD}_{\text{2.0}}}\) & \textbf{0.60} & \underline{0.29} & \underline{12.83} & 15.01 & 32.72 & \underline{6.56} & 44.17 \\
        \(\text{SDXL}_{\text{1.0}}\) & \underline{0.61} & \underline{0.29} & \textbf{11.85} & \textbf{16.01} & \textbf{33.54} & 6.50 & / \\
		MD\(_{\Phi=\text{SDXL}_{\text{1.0}}}\) & 0.67 & \textbf{0.28} & \textbf{11.85} & 14.74 & 32.40 & 6.37 & 172.62 \\
        Ours\(_{\Phi=\text{SDXL}_{\text{1.0}}}\) & \textbf{0.60} &\textbf{0.28} & \textbf{11.85} & \underline{15.63} & \underline{32.78} & \textbf{6.58} & 174.46 \\
		\hline
	\end{tabular}}
\end{table}

\end{document}